\documentclass[times,review,10pt]{elsarticle}

\usepackage[hyphens]{url}
\usepackage{graphicx}
\usepackage{booktabs}
\usepackage{amsmath}
\usepackage{amssymb}
\usepackage{tabularx}
\usepackage{makecell}
\usepackage{multirow}
\usepackage{float}
\usepackage{xcolor}
\usepackage[a4paper,top=4.3cm,right=4.8cm,bottom=4.3cm,left=4.8cm]{geometry}

\journal{Pattern Recognition}

\begin{document}

\begin{frontmatter}

\title{\fontsize{14}{17}\selectfont CARNet: End-to-End Coupled Association-Recognition Network for Fine-Grained Multi-Drone Collaborative Detection}
\author[aff1]{\fontsize{8}{10}\selectfont Zhao Zhang}
\author[aff1]{\fontsize{8}{10}\selectfont Ruigang Fu\corref{cor1}}
\ead{furuigang08@nudt.edu.cn}
\author[aff1]{\fontsize{8}{10}\selectfont Yu Zhang}
\author[aff1]{\fontsize{8}{10}\selectfont Jiajun Xiang}
\author[aff1]{\fontsize{8}{10}\selectfont Zixuan Xiao}
\author[aff1]{\fontsize{8}{10}\selectfont Ping Zhong}
\cortext[cor1]{Corresponding author}
\address[aff1]{College of Electronic Science and Technology, National University of Defense Technology, Changsha, China}

\begin{abstract}
Multi-drone collaborative detection aims to enhance target recognition by leveraging complementary observations from overlapping aerial viewpoints. Existing approaches typically follow a two-stage pipeline that first establishes cross-view correspondences and then performs fusion-based recognition. However, this decomposition suffers from inconsistent optimization objectives, redundant pairwise modeling, and irreversible error propagation, where incorrect associations inevitably undermine subsequent recognition performance. To overcome these limitations, we propose the Coupled Association-Recognition Network (CARNet), a unified framework that jointly optimizes cross-view target association and recognition fusion through a shared target-pair representation. To improve association reliability in densely populated scenes with visually similar targets, CARNet explicitly incorporates spatial cues into pairwise relation modeling. Moreover, we introduce two complementary causal-inspired intervention strategies: the first enhances cross-view association by emphasizing appearance-confusing pairs, while the second bolsters the robustness of fusion-based recognition by learning stable multi-view representations under incomplete visual evidence. In addition, we present MDMT-CD, a fine-grained extension of the MDMT dataset that enriches the original vehicle annotations with detailed category labels, providing the first benchmark specifically designed for multi-drone collaborative vehicle detection. Extensive experiments demonstrate that CARNet consistently outperforms existing state-of-the-art methods.  The MDMT-CD dataset and source code will be available at \url{https://github.com/bigdreaming11/CARNet}.
\end{abstract}


\end{frontmatter}

\section{Introduction}

Single-drone object detection is prone to classification errors and struggles to handle challenging scenarios such as occlusion \cite{liu2026spformer}. In contrast, multi-drone systems enable collaborative observation of the same target from different aerial viewpoints, providing complementary structural and appearance information \cite{liu2026topological}.



A natural implementation is a two-stage pipeline: an association network first establishes correspondences, and a subsequent fusion classifier recognizes each matched target \cite{peng2023skynet}. However, the two stages are optimized with heterogeneous objectives, leading to inconsistencies in representation learning and redundant pairwise modeling across modules. Moreover, the cascade design introduces a structural bottleneck where association errors are irrevocably propagated to the recognition stage \cite{lu2026skyseg}. Once the first stage incorrectly links two different vehicles, subsequent feature fusion inevitably aggregates conflicting evidence, making the final prediction unrecoverable. This design is also computationally redundant and inefficient due to duplicated pairwise modeling across stages. This motivates a unified formulation in which target association and fusion recognition are learned through the same target-pair representation.


We address this bottleneck with a position-aware all-pairs relation. For every target in one drone view and every target in the other, we construct a fused relation by combining their cross-view visual descriptors.  LGSGMM \cite{zhao2026two}demonstrat , positional information provides valuable geometric cues that improve feature discriminability and correspondence estimation under appearance ambiguity. Motivated by this observation, we encode each target center using a learnable Fourier embedding and inject the resulting positional representation into the fused relation, allowing it to jointly model visual appearance and spatial consistency across views. Each fused target-pair relation is mapped to a $C+1$ distribution containing $C$ fine-grained vehicle categories and one no-match state. Consequently, the relation tensor serves as a unified target-pair representation that jointly performs cross-view target association and fusing recognition within a single prediction stage, eliminating the conventional separation between association and fusion-based recognition. This unified formulation enables CARNet to replace the conventional association network and fusion recognition with a single prediction stage.

However, even with the unified target-pair modeling scheme, association learning may still struggle to capture discriminative correspondence evidence from challenging target pairs. Existing supervision paradigms typically optimize pairwise relations based on matching labels, while overlooking the varying discriminative difficulty among different target pairs. Since easy negative pairs provide straightforward optimization signals, the model tends to prioritize learning easily separable appearance patterns, whereas the fine-grained identity cues and critical correspondence evidence embedded in highly similar target pairs remain insufficiently explored, resulting in incorrect associations in complex scenarios.

Motivated by Pearl’s causal hierarchy\cite{guyon2010causality}, CARNet introduces two complementary intervention strategies to guide the model toward learning stable cross-view correspondence evidence. The first enhances cross-view correspondence by assigning greater learning emphasis to appearance-confusing non-matching target pairs, encouraging the model to devote more optimization capacity to challenging correspondence decisions. The second intervention perturbs the visual evidence within target pairs and enforces consistency between the original and intervened relation representations, encouraging the model to learn stable correspondence representations that remain robust under visual variations and incomplete observations.
%
%
%

Our contributions are summarized as follows:
\begin{itemize}
	
\item We establish MDMT-CD, a fine-grained extension of the MDMT dataset by enriching the original single-category vehicle annotations with fine-grained category labels.
	\item We propose CARNet, a unified pairwise relation framework that integrates cross-view correspondence and fusion recognition in a single prediction stage. 
	
	\item 	We propose two causal-inspired intervention strategies that improve cross-view correspondence in densely distributed scenes with appearance-similar targets while enhancing the robustness of multi-view fusion recognition.
	
	\item CARNet achieves state-of-the-art collaborative detection and cross-view association performance on the MDMT-CD  datasets, the best association performance on the infrared and cross-modal subsets of M3Track, and further demonstrates excellent generalization on the cross-camera multi-object tracking benchmarks Wildtrack and MvMHAT.

\end{itemize}
\section{Related Work}

\subsection{Multi-Drone Collaborative Perception}

Multi-drone collaborative perception exploits complementary observations from multiple UAVs to alleviate limited fields of view, occlusion, small object scales, and unstable imaging conditions. Early studies mainly focused on collaborative tracking. Zhu et al.~\cite{zhu2021multidrone} proposed an agent-sharing network that exchanges reliable target templates and performs view-aware fusion for multi-drone single-object tracking. Liu et al.~\cite{liu2023robust} introduced the MDMT benchmark and MIA-Net, which combine within-view tracking with cross-drone identity authentication to recover targets occluded in individual views. Wu et al.~\cite{wu2025temporal} further enhanced multi-drone multi-object tracking through temporal--spatial feature interaction. Beyond UAV-specific settings, Nguyen et al.~\cite{nguyen2024multi} formulated multi-camera multi-object tracking as a single-stage global association problem, jointly reasoning about detections and cross-camera correspondences in overlapping views. Zheng et al.~\cite{zheng2024motion} further improved multi-object association under occlusion through motion-guided cross-temporal aggregation and hierarchical matching.

Recent research has increasingly investigated feature-level collaborative detection. DiscoNet~\cite{li2021disconet} learns a collaboration graph under teacher--student distillation, whereas Where2comm~\cite{hu2022where2comm} communicates only spatial regions with high perceptual confidence. V2X-ViT~\cite{xu2022v2xvit}, CoBEVT~\cite{xu2023cobevt} and CoBEVFlow~\cite{wei2023cobevflow} further address heterogeneous observations, sparse cross-agent fusion, communication redundancy, and temporal asynchrony. Although several of these methods were initially developed for vehicle-based agents, their feature alignment and communication mechanisms are also applicable to UAV collaboration.

For UAV-specific three-dimensional perception, UAV3D~\cite{ye2024uav3d} provides benchmarks for collaborative 3D detection and tracking, while UCDNet~\cite{tian2025ucdnet} maps multi-UAV image features into a common 3D space using depth priors and geometric consistency. MCOP~\cite{lin2025mcop} extends collaboration from bounding-box detection to dense occupancy prediction with altitude-aware compression. V2U4Real~\cite{li2026v2u4real} studies real-world vehicle-to-UAV cooperative detection, and AirCopBench~\cite{zha2026aircopbench} evaluates multi-drone embodied perception and reasoning. Nevertheless, most existing methods output trajectories, boxes, or occupancy states, while target-level correspondence and fine-grained collaborative recognition remain insufficiently coupled.

\subsection{Cross-View Target Association}

Cross-view target association aims to determine whether targets observed by different UAVs correspond to the same physical object. Compared with conventional multi-object tracking, the task is considerably more challenging due to drastic viewpoint changes, severe rotations, scale variations, target occlusions, and dense distributions of visually similar objects.

Existing UAV-oriented methods mainly exploit geometric consistency to establish cross-view correspondences. Li \emph{et al.}~\cite{li2022triangular} represented each target using triangular topological sequences and performed globally consistent association through two-stage assignment. Liu \emph{et al.}~\cite{liu2023mdmt} introduced the MDMT benchmark and proposed MIA-Net, which first establishes image-level correspondences using SIFT features and then associates tracked targets across UAVs through geometric verification. Qiao \emph{et al.}~\cite{qiao2024stca} proposed STCA, which estimates inter-view transformations, identifies common visible regions, and performs bidirectional cross-drone association. Tan \emph{et al.}~\cite{tan2025gtanet} further formulated target association as a graph matching problem and exploited reliable spatial relationships to improve robustness under missed and false detections.

Besides these dedicated UAV-oriented methods, appearance-based ReID models and deep image matching networks are also widely adopted as association baselines. However, appearance-based methods such as Re-ID \cite{he2023fastreid}, OSNet \cite{zhou2019omni}, ReIDMamba \cite{gu2026reidmamba},  CrossMOT \cite{hao2024divotrack}, CT \cite{wieczorek2021unreasonable}, AGW \cite{ye2021deep}, mainly rely on visual similarity and therefore struggle to distinguish densely distributed targets with highly similar appearances. Deep image matching methods, such as SuperGlue~\cite{sarlin2020superglue}, LoFTR~\cite{sun2021loftr} and Roma~\cite{edstedt2024roma}, often experience noticeable performance degradation under the large rotation differences commonly encountered between UAV views. Moreover, MIA-Net depends on handcrafted SIFT features for image registration, resulting in relatively low computational efficiency. Furthermore, most existing methods perform target association as an independent preprocessing stage, making it difficult to jointly optimize cross-view association and downstream collaborative recognition.

\section{Fine-Grained MDMT Annotation}
The original MDMT dataset assigns all vehicle targets to a single category, car, which limits its capability for fine-grained semantic analysis. In this work, we extend MDMT by retaining the original cross-view identity annotations and further refining the object labels into seven fine-grained categories: car, bus, van, truck, tricycle, bicycle, and pedestrian. This extension enables more comprehensive evaluation of multi-view target perception and association under diverse object types.  
In addition, we manually verify the entire dataset and correct annotation errors from the original release to improve annotation reliability. Each annotation record consists of the object category, target identity (ID), and bounding box information.

\section{Method}
\begin{figure}[t]
	\centering
	\includegraphics[width=1\columnwidth]{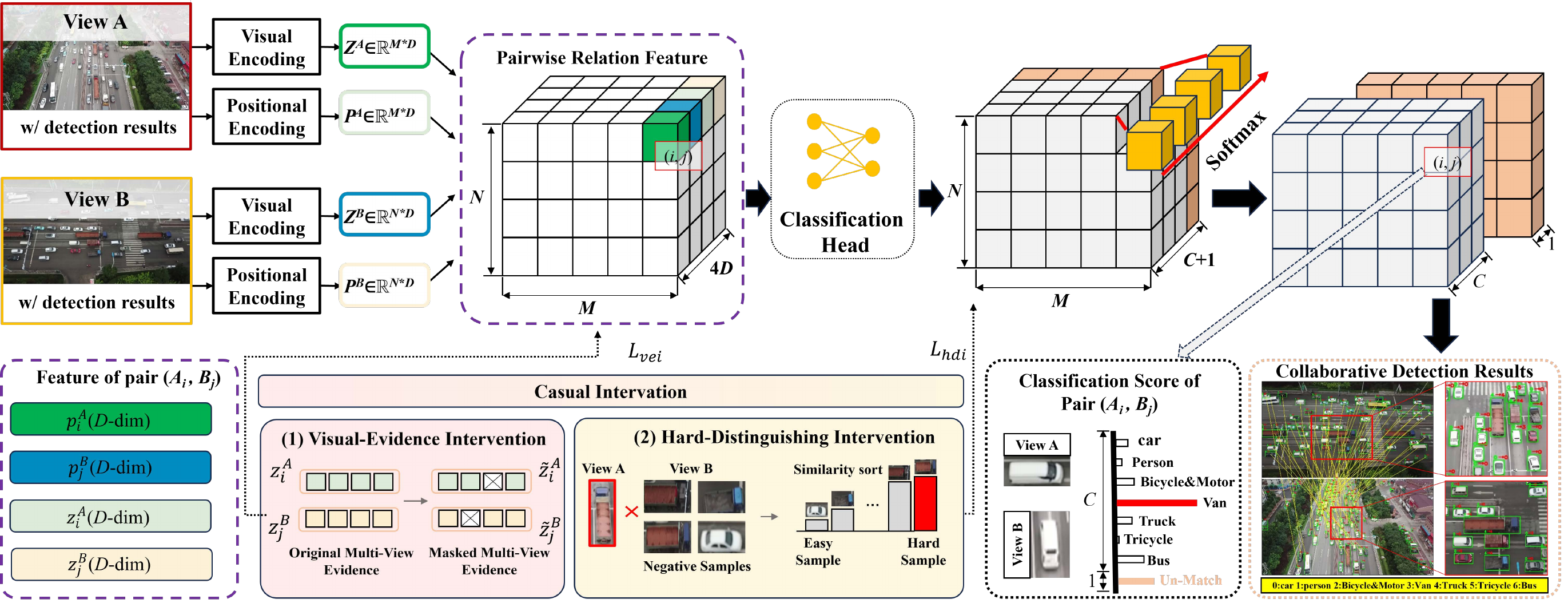}
	\caption{Overall pipeline of the proposed CARNet. Given two drone views, target features are extracted and used to construct position-aware all-pairs relations. Each target-pair relation is mapped to a unified $C+1$ distribution for joint cross-view association and collaborative recognition, while two causal-inspired intervention  are applied  during training.}
	\label{流程图}
\end{figure}
Given a pair of UAV images $(I^A,I^B)$, the onboard object detector first extracts
$M$ target instances from $I^A$ and $N$ target instances from $I^B$.
Conventional collaborative detection methods generally decompose the task into two successive stages. First, cross-view object association is performed to establish correspondences between targets observed from different UAVs. Then, multi-view feature fusion is conducted for the matched target pairs to obtain the final fused recognition results.

In contrast, we formulate object association and multi-view feature fusion as a unified end-to-end framework, enabling one-stage collaborative object detection for multi-drone systems. The overall pipeline of the proposed method is illustrated in Fig.~\ref{流程图}.

\subsection{Position-Aware Pairwise Relation Modeling}
Fistly, a shared backbone is employed to process both UAV images. Subsequently, ROI pooling extracts one feature vector for each detected target, resulting in the object feature sets $Z^A\in\mathbb{R}^{M\times D}$,$Z^B\in\mathbb{R}^{N\times D}$.
When visually similar targets are densely distributed, appearance features alone are often insufficient to distinguish different instances. To alleviate this ambiguity, we introduce positional information to facilitate cross-view association. Specifically, following the use of learned positional encodings in modern feature matching \citep{sarlin2020superglue}, the center coordinates of each target are mapped by a learnable Fourier encoder, yielding a positional feature
$p\in\mathbb{R}^{D}$ which has the same dimensionality as the visual descriptor. For each cross-view candidate pair $(i,j)$, we construct a position-aware pairwise relation representation by jointly encoding the appearance descriptors and positional information of the object pair:
\begin{equation}
	r_{ij}=\left[
	z_i^A,\ z_j^B,\ p_i^A, \ p_j^B
	\right]
	\label{eq:relation}
\end{equation}

 Together with the Fourier positional difference, the relation representation supplies complementary evidence for association and collaborative detection. Thus, the pairwise relation matrix between the two views is obtained as:
\begin{equation}
r\in\mathbb{R}^{ M\times N\times4D}
\end{equation}

An MLP maps each relation feature $r_{ij}$ to a $(C+1)$-dimensional relation distribution:
\begin{equation}
\pi_{ij}
=
[\pi_{ij,1},\ldots,\pi_{ij,C},\pi_{ij,\varnothing}].
\label{eq:relation-distribution}
\end{equation}

Here, $C$ is the number of fine-grained target categories. $\pi_{ij,c}$ denotes the probability that pair $(i,j)$ is a valid correspondence with category $c$, where $c\in\{1,\ldots,C\}$. $\pi_{ij,\varnothing}$ denotes the probability that pair $(i,j)$ belongs to the \emph{no-match} class. The final decoder uses a fast bidirectional best-match strategy to determine valid correspondences. Only accepted target pairs are used to update the single-view detection categories.

\subsection{Hard-Distinguishing Intervention}

Dense scenes contain many visually similar vehicles. This may create an appearance shortcut: visually similar targets are treated as matched targets. Although the pairwise loss supervises all valid target pairs, the optimization contribution of difficult relations remains limited because the relation distribution is highly imbalanced and dominated by easy no-match pairs. To suppress this shortcut, we introduce a hard non-match intervention. This intervention does not generate new training pairs. Instead, it selects appearance-confusing no-match pairs from existing all-pairs relations and imposes additional no-match supervision on them.

Let $\mathcal{P}^{+}$ denote the set of ground-truth matched target pairs. For each target $i$ in view A, we construct a candidate non-match set by excluding its ground-truth correspondence:

\begin{equation}
	\mathcal{N}_i=
	\{j\mid (i,j)\notin \mathcal{P}^{+}\}.
	\label{eq:nonmatch-set}
\end{equation}
Here, $j$ indexes a target in view B. $\mathcal{N}_i$ contains all valid targets in view B that should not be matched with target $i$.

The visual descriptors are first normalized. Then, the most confusing non-match target is selected according to cosine similarity:

\begin{equation}
	j_i^-=
	\arg\max_{j\in\mathcal{N}_i}
	\frac{z_i^A\cdot z_j^B}
	{\|z_i^A\|_2\|z_j^B\|_2}.
	\label{eq:hard-negative}
\end{equation}
where $\cdot$ denotes inner product and $\|\cdot\|_2$ denotes the $\ell_2$ norm. The selected target $j_i^-$ is the no-match target that is most similar to $i$ in the descriptor space.

The intervention set is defined as:

\begin{equation}
	\mathcal{H}
	=
	\{(i,j_i^-)\mid \mathcal{N}_i\neq\emptyset\}.
	\label{eq:hard-set}
\end{equation}

Where $\mathcal{H}$ contains the selected hard non-match pairs. Each pair in $\mathcal{H}$ is a true negative pair, but it has high visual similarity. For these intervened pairs,  the model is encouraged to classify them as unmatched:

\begin{equation}
	\mathcal{L}_{\mathrm{hdi}}
	=
	-\frac{1}{|\mathcal{H}|}
	\sum_{(i,j)\in\mathcal{H}}
	\log \pi_{ij,\varnothing}.
	\label{eq:hnm-loss}
\end{equation}

Where $|\mathcal{H}|$ is the number of selected Hard-Distinguishing pairs. The selected pairs in $\mathcal{H}$ are already included in $\mathcal{V}$ and are therefore also supervised by $\mathcal{L}_{\mathrm{pair}}$. The role of $\mathcal{L}_{\mathrm{hdi}}$ is to increase the optimization weight of these Hard-Distinguishing pairs.

\subsection{Visual-Evidence Perturbation Intervention}

The second intervention acts on visual evidence. Cross-view observations are often incomplete because of viewpoint changes, occlusion, illumination variation, and scale changes. To simulate this condition, independent dropout masks are applied to target descriptors:

\begin{equation}
	\widetilde z_i^A=m_i^A\odot z_i^A,
	\quad
	\widetilde z_j^B=m_j^B\odot z_j^B .
	\label{eq:vei-dropout}
\end{equation}

Where $m_i^A$ and $m_j^B$ are independently sampled dropout masks for the two views. $\widetilde z_i^A$ and $\widetilde z_j^B$ are the perturbed descriptors after partial removal of visual evidence.

The intervened relation representation $\widetilde r_{ij}$ is reconstructed using the same relation module as the clean relation $r_{ij}$. The position embeddings are not dropped. Thus, the intervention is restricted to the visual descriptors. A consistency constraint is imposed between the original and intervened relations:

\begin{equation}
	\mathcal{L}_{\mathrm{vei}}
	=
	\frac{1}{|\mathcal{V}|}
	\sum_{(i,j)\in\mathcal{V}}
	\|
	\widetilde r_{ij}
	-
	\operatorname{sg}(r_{ij})
	\|_2^2 .
	\label{eq:vei-loss}
\end{equation}
where $\mathcal{V}$ denotes the set of valid target pairs used for relation consistency, and $\operatorname{sg}(\cdot)$ denotes the stop-gradient operation. The clean relation is treated as a fixed target. The consistency constraint encourages the relation representation to preserve reliable correspondence information under visual perturbations.
\subsection{Training Objective}
The basic supervision is imposed on all valid target pairs. For each pair $(i,j)\in\mathcal{V}$, we define its relation label as:
\begin{equation}
	y_{ij}=
	\begin{cases}
		c_i, & (i,j)\in\mathcal{P}^{+},\\
		\varnothing, & (i,j)\notin\mathcal{P}^{+},
	\end{cases}
	\label{eq:pair-label}
\end{equation}
where $c_i\in\{1,\ldots,C\}$ is the fine-grained category label of target $i$. Thus, a positive pair is supervised by its fine-grained category, while an unmatched pair is supervised as no-match.
The all-pairs relation classification loss is defined as
\begin{equation}
	\mathcal{L}_{\mathrm{pair}}
	=
	-\frac{1}{|\mathcal{V}|}
	\sum_{(i,j)\in\mathcal{V}}
	\log \pi_{ij,y_{ij}} .
	\label{eq:pair-loss}
\end{equation}
 This loss optimizes all valid target pairs, where matched pairs are supervised by fine-grained categories and unmatched pairs are supervised by the no-match state.

The overall objective is

\begin{equation}
	\mathcal{L}
	=
	\mathcal{L}_{\mathrm{pair}}
	+
	\lambda_1\mathcal{L}_{\mathrm{hdi}}
	+
	\lambda_2\mathcal{L}_{\mathrm{vei}}.
	\label{eq:total-loss}
\end{equation}
where $\lambda_1$ and $\lambda_2$ are learnable scalar parameters that control the contributions of the two intervention losses. After training, $\lambda_1$ and $\lambda_2$ converge to approximately 0.027 and 0.042, respectively.

\section{Experiments}

\subsection{Datasets and Experimental Settings}
We evaluate the proposed method on four benchmarks. MDMT-CD is used for multi-drone collaborative detection and target association, while  M3Track are used for cross-view target association. To further examine generalization, we also conduct experiments on two cross-camera multi-person tracking benchmarks, Wildtrack and MvMHAT.

MDMT-CD contains small-scale aerial targets, dense appearance-similar objects, severe occlusions, and large viewpoint variations, making cross-view association challenging. M3Track contains low-altitude UAV and ground-camera views, including both visible-spectrum and infrared sequences, and is used to evaluate association under modality and viewpoint changes. The image resolutions of MDMT and M3Track are 1920$\times$1080 and 640$\times$512, respectively.

For collaborative detection, we report mAP@50 and mAP@50--90. For association on MDMT-CD and M3Track, we follow the MDMT protocol and report MDA. For Wildtrack and MvMHAT, we report MOTA and IDF1. All experiments are conducted on two NVIDIA RTX 4090 GPUs with a batch size of 4. YOLOv8 is used as the single-drone detector. For cross-camera tracking experiments, we replace the target association module of MIA-Net with CARNet while keeping the remaining components unchanged.
In the association experiments, we set the number of categories $C$ to 1.

\begin{table}[ht]
	\centering
	{
		\setlength{\tabcolsep}{2pt}
		\begin{tabular}{lccc}
			\toprule
			Method  & $mAP_{50}$ & $mAP_{50-90}$ & Latency (ms)\\
			\midrule
			Single Drone             & 23.45 & 16.12 & -- \\
			UVF (AAAI 2026)          &16.40 & 11.00 & 173 \\
			MVCNN (CVPR 2015)        & 17.39 & 11.55 & 88 \\
			CMLP                     & 17.96 & 12.09 & 103 \\
			SMDC (AAAI 2023)         & 20.99 & 14.19 & 94 \\
			ARA (AAAI 2025)          & 21.31 & 14.41 & 97 \\
			DS (IGRSL 2015)          & 22.43 & 15.15 & 86 \\
			DC                		 & 23.71 & 15.89 & 83 \\
			TMC (TPAMI 2023)         & 23.67 & 16.33 & 89 \\
			\textbf{CARNet(Ours)}                    & \textbf{25.23} & \textbf{17.21} & \textbf{48} \\
			\bottomrule
		\end{tabular}
	}
	
	\caption{Comparison of collaborative detection methods on the MDMT-CD dataset. The best scores are marked in bold.}
	\label{Comparison1}
\end{table}

\begin{figure}[htbp]
	\centering
	\includegraphics[width=0.85\columnwidth]{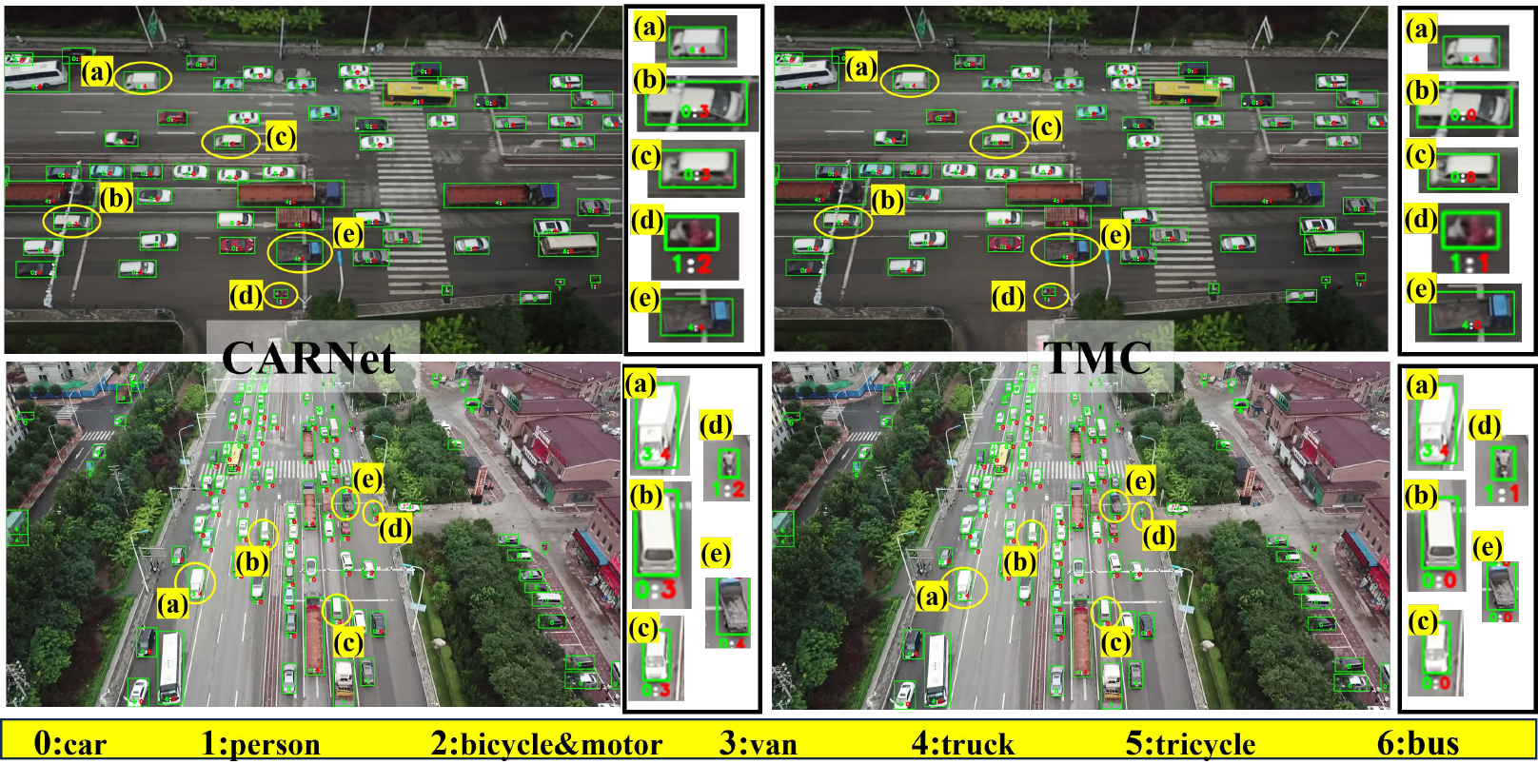}
	\caption{Visualization of collaborative detection results on the  MDMT-CD dataset. Green numbers denote single-drone predictions, and red numbers denote fused predictions. Boxes without red numbers are not associated.}
	\label{识别图}
\end{figure}
\subsection{Comparison of Collaborative Detection Performance}

Since there are currently no mature methods specifically designed for collaborative object detection in multi-drone systems, we construct a strong baseline by combining the state-of-the-art cross-view object association method LGSGMM \cite{zhao2026two} with representative multi-view classification models, forming an "target association--fusing recognition" pipeline. Specifically, the compared fusion methods include DS \cite{dong2015target}, MVCNN \cite{su2015multi}, TMC \cite{han2022trusted}, SMDC \cite{liu2023safe}, ARA \cite{liu2025enhancing}, and UVF \cite{lv2026uncertainty}. In addition, we consider two simple yet widely adopted engineering baselines. The first selects the final category according to the detection confidence of a single drone, referred to as DC (Decision by Confidence). The second directly concatenates the features extracted from different views and feeds them into an MLP for classification, referred to as CMLP (Concatenation + MLP). To evaluate the effectiveness of CARNet, we compare it with the aforementioned single-view, decision-level fusion, and feature-level fusion approaches. The quantitative results are reported in Table~\ref{Comparison1}, while qualitative comparisons are presented in Fig.~\ref{识别图}.

As shown in Table~\ref{Comparison1}, decision-level fusion methods based on single-drone logits or class probabilities, such as DS, TMC, SMDC, and ARA, consistently outperform visual feature fusion methods such as MVCNN and UVF. This is because category outputs directly provide task-relevant discriminative evidence, while evidence modeling, uncertainty estimation, and unreliable-view rejection further alleviate cross-view prediction conflicts, thereby improving multi-drone collaborative recognition accuracy. However, these methods do not fully exploit the visual and positional information of the targets. CARNet further models all $M\times N$ cross-view target pairs through a unified $C+1$ relation distribution. By jointly exploiting visual interactions, spatial cues, and intervention-based training, it improves association among densely distributed similar vehicles and performs association, fusion, and recognition within a single framework. Compared with the single-drone baseline, CARNet improves $mAP_{50}$ by 1.78\% $(25.23\%-23.45\%)$. Compared with the strongest existing collaborative detection baseline, TMC, CARNet improves $mAP_{50}$ by 1.56\% $(25.23\%-23.67\%)$ . Meanwhile, CARNet reduces the latency from 89 ms to 48 ms, demonstrating both higher accuracy and better inference efficiency.

\begin{table}[t]
	\centering
	{
		\setlength{\tabcolsep}{3pt}
		\begin{tabular}{lcccc}
			\toprule
			Method & \makecell{Precision\\(\%)} & \makecell{Recall\\(\%)} & \makecell{MDA\\(\%)} & \makecell{Latency\\(ms)}\\
			\midrule
			ReIDMamba (TMM 2026)      & 30.07 & 28.96 & 14.22 & 107 \\
			Re-ID (ACM MM 2023)       & 31.53 & 28.27 & 15.31 & \textbf{18 }\\
			OSNet (ICCV 2019)         & 33.11 & 30.13 & 18.11 & 21 \\
			TTS (Drones 2022)         & 26.97 & 24.57 & 12.06 & 1284 \\
			GTA-Net (Drones 2025)     & 34.64 & 32.31 & 19.27 & 1034 \\
			MambaGlue (CVPR 2025)     & 8.27  & 8.64  & 2.87  & 44 \\
			SuperGlue (CVPR 2020)     & 9.94  & 8.09  & 3.05  & 38 \\
			LoFTR (CVPR 2021)         & 11.24 & 10.01 & 4.07  & 625 \\
			E-LoFTR (CVPR 2024)       & 10.87 & 10.01 & 3.66  & 517 \\
			JamMa (CVPR 2025)         & 16.77 & 14.21 & 5.94  & 633 \\
			RoMa (CVPR 2024)          & 18.31 & 12.68 & 6.73  & 1176 \\
			MIA-Net (TMM 2023)        & 47.60 & 48.81 & 36.88 & 820 \\
			LGSGMM (TCSVT 2026)       & 54.17 & \textbf{56.11} & 39.29 & 80 \\
			\textbf{CARNet(Ours)}  			         &  \textbf{56.67}& 49.22 & \textbf{40.81}&48  \\
			\bottomrule
		\end{tabular}
	}
	\caption{Comparison of cross-view object association methods on the MDMT-CD dataset. The best scores are marked in bold.}
	\label{mdmt}
\end{table}

\subsection{Comparison of Cross-View Object Association Performance}
\subsubsection{Comparison on MDMT-CD}

\begin{figure}[t]
	\centering
	\includegraphics[width=0.9\columnwidth]{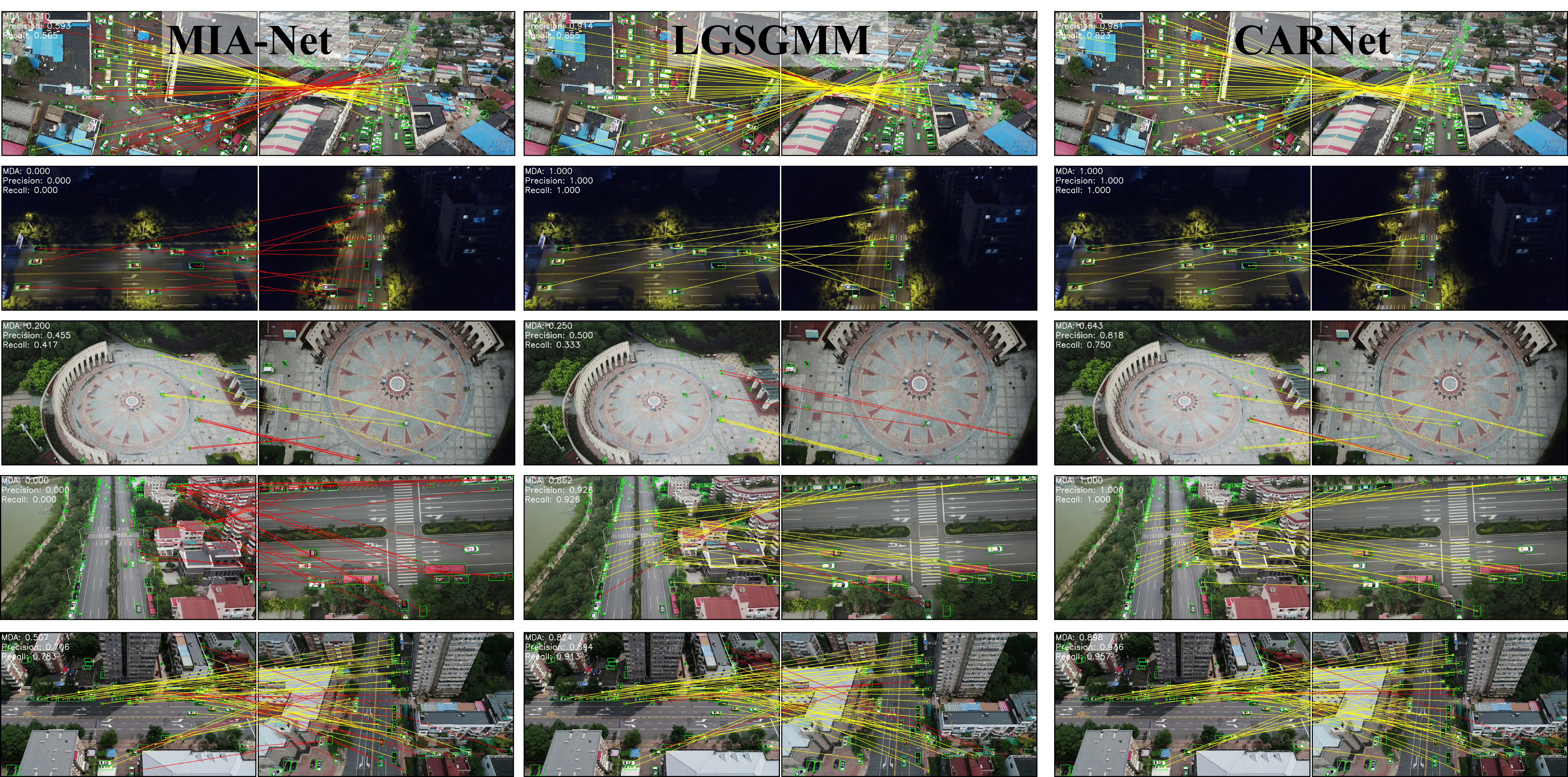}
	\caption{Visualization of cross-view target association results on the MDMT-CD dataset. Red lines indicate incorrect associations, while yellow lines denote correct associations.}
	\label{m3_pic}
\end{figure}
To further evaluate the effectiveness of the proposed method in scenarios with densely distributed similar targets, we compare it with representative state-of-the-art cross-view object association methods on the MDMT-CD dataset, including Re-ID , OSNet , ReIDMamba , GTA-Net, TTS \cite{li2022multi}, MIA-Net , SuperGlue , LoFTR , E-LoFTR , JamMa, MambaGlue \cite{ryoo2025mambaglue}, and RoMa . Among them, MIA-Net is based on SIFT, while SuperGlue, LoFTR, E-LoFTR, JamMa, MambaGlue, and RoMa are state-of-the-art deep learning-based image matching methods that estimate homography matrices for cross-view object mapping. The quantitative results are reported in Table~\ref{mdmt}. As shown in the table, image matching-based object association methods generally struggle with large rotation variations between views, resulting in the poorest performance on the MDMT-CD dataset. Appearance similarity-based association methods are also prone to incorrect associations due to the presence of numerous visually similar objects in UAV views. In contrast, the proposed CARNet achieves a 3.93\% improvement over MIA-Net in terms of MDA (40.81\% vs. 36.88\%) while maintaining more than 10× higher computational efficiency. Compared with the current state-of-the-art LGSGMM, CARNet achieves comparable association performance with significantly improved efficiency.
Fig.~\ref{m3_pic} shows the qualitative association results. As can be observed, CARNet can effectively associate densely distributed and visually similar objects, even under large rotation and viewpoint variations between image pairs. Moreover, it maintains robust association performance in nighttime scenarios with sparse object appearances. This improvement benefits from the incorporation of positional information and the proposed full object-pair modeling strategy, which jointly exploit spatial and visual relationships for discriminative association.

\begin{table}[p]
	
	\centering
	
	{
		\setlength{\tabcolsep}{2pt}
		\renewcommand{\arraystretch}{0.65}
		\begin{tabular}{l c c c c}
			\Xhline{2\arrayrulewidth}
			& Method & \makecell{Precision\\(\%)} & \makecell{Recall\\(\%)} & \makecell{MDA\\(\%)} \\
			\Xhline{2\arrayrulewidth}
			
			\multirow{10}{*}{\textbf{Visible Light}}
			
			& TTS {\selectfont (Drones 2022)} & 9.89 & 18.21 & 3.48 \\
			& GTA-Net {\selectfont (Drones 2025)} & 16.78 & 13.23 & 5.46 \\
			& MambaGlue {\selectfont (CVPR 2025)} & 17.31 & 14.21 & 6.33 \\
			& MIA-Net {\selectfont (TMM 2023)} & 29.77 & 33.21 & 15.67 \\
			& ReIDMamba {\selectfont (TMM 2026)} &39.99 & 43.23 & 19.86 \\
			& OSNet {\selectfont (ICCV 2019)} & 40.21 & 46.32 & 20.76 \\
			& CrossMOT {\selectfont (IJCV 2024)} & 39.77 & 49.36 & 21.07\\
			& CT {\selectfont (ICONIP 2021)} & 42.48 & 54.50 & 24.85\\
			& AGW {\selectfont (TPAMI 2021)} & 43.42 & 56.40 & 27.47 \\
			& LGSGMM {\selectfont (TCSVT 2026)} & 46.05 &\textbf{46.38} & 30.05 \\
			& 			\textbf{CARNet(Ours)}    &\textbf{66.14 } &40.28 &\textbf{33.40}  \\
			
			\hline\hline
			
			\multirow{10}{*}{\textbf{Infrared}}
			& TTS {\selectfont (Drones 2022)} & 3.04 & 16.78& 4.01 \\
			& GTA-Net {\selectfont (Drones 2025)} & 7.81 & 12.96& 4.61 \\
			& ReIDMamba {\selectfont (TMM 2026)} & 25.47 & 32.11 & 10.24\\
			& OSNet {\selectfont (ICCV 2019)} & 27.87 & 35.28 & 12.88 \\
			& MambaGlue {\selectfont (CVPR 2025)} & 26.34 & 36.42 & 12.97 \\
			& CrossMOT {\selectfont (IJCV 2024)} & 24.36 & 42.99 & 13.20 \\
			& MIA-Net {\selectfont (TMM 2023)} & 28.03& 39.77 & 14.22 \\
			& CT {\selectfont (ICONIP 2021)} & 28.82& 37.25 & 14.62 \\
			& AGW {\selectfont (TPAMI 2021)} & 39.58 & 51.16 & 22.54 \\			
			& LGSGMM {\selectfont (TCSVT 2026)} &41.67 &\textbf{37.49}& 24.59\\
			& 			\textbf{CARNet(Ours)}     & \textbf{61.41}&36.84 &\textbf{29.91} \\
			
			
			\hline\hline
			
			\multirow{9}{*}{\textbf{Cross Modal}}	
			& GTA-Net {\selectfont (Drones 2025)} & 7.66 & 11.29& 4.36 \\
			& ReIDMamba {\selectfont (TMM 2026)} & 15.03 & 16.23 & 5.31\\
			& CT {\selectfont (ICONIP 2021)} & 14.04& 18.13 & 6.19\\
			& AGW {\selectfont (TPAMI 2021)} & 14.47 & 18.68 & 6.41 \\			
			& MIA-Net {\selectfont (TMM 2023)} & 18.36& 12.29 & 8.74 \\
			& DASF {\selectfont (TMM 2025)} &27.24&18.76& 11.32\\	
			& DLM {\selectfont (TPAMI 2025)} &26.32&20.78& 12.06\\	
			& LAReVIT
			{\selectfont (CVPR 2026)} &29.18&23.07& 14.79\\
			& LGSGMM {\selectfont (TCSVT 2026)} &28.94&\textbf{26.62}& 16.10\\
			& 			\textbf{CARNet(Ours)}     & \textbf{42.89}&21.03 &\textbf{16.43} \\
			
			\Xhline{2\arrayrulewidth}
		\end{tabular}
	}
	\caption{Comparison of Different Methods on M3Track dataset. The best scores are marked in bold. }
	\label{tab:m3}
\end{table}

\begin{figure}[t]
	\centering
\includegraphics[width=0.9\columnwidth]{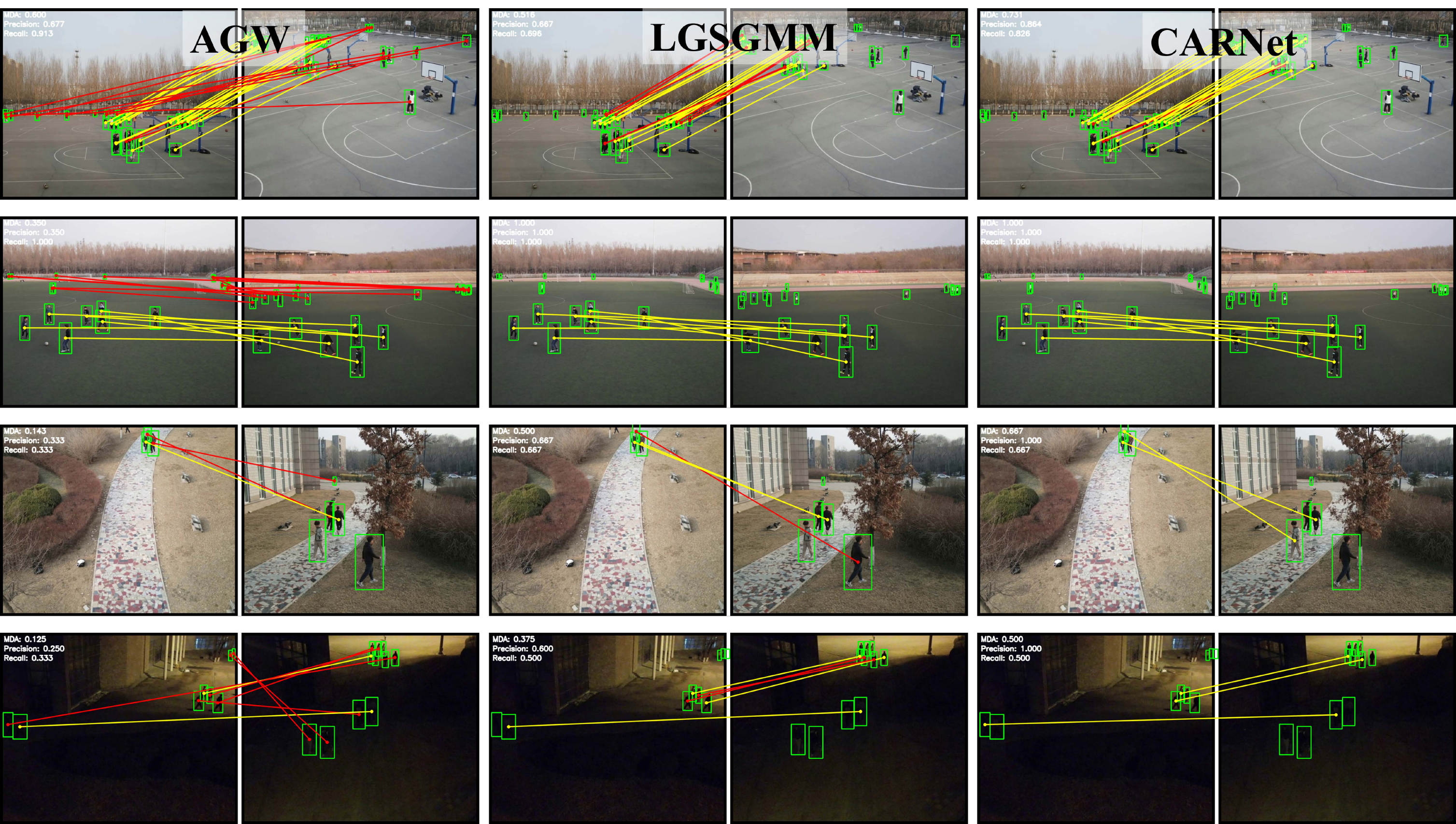}
\caption{Visualization of cross-view target association results on the visible-light subset of the M3Track dataset.}
\label{m3可见光}

\end{figure}

\begin{figure}[t]
	\centering
\includegraphics[width=0.9\columnwidth]{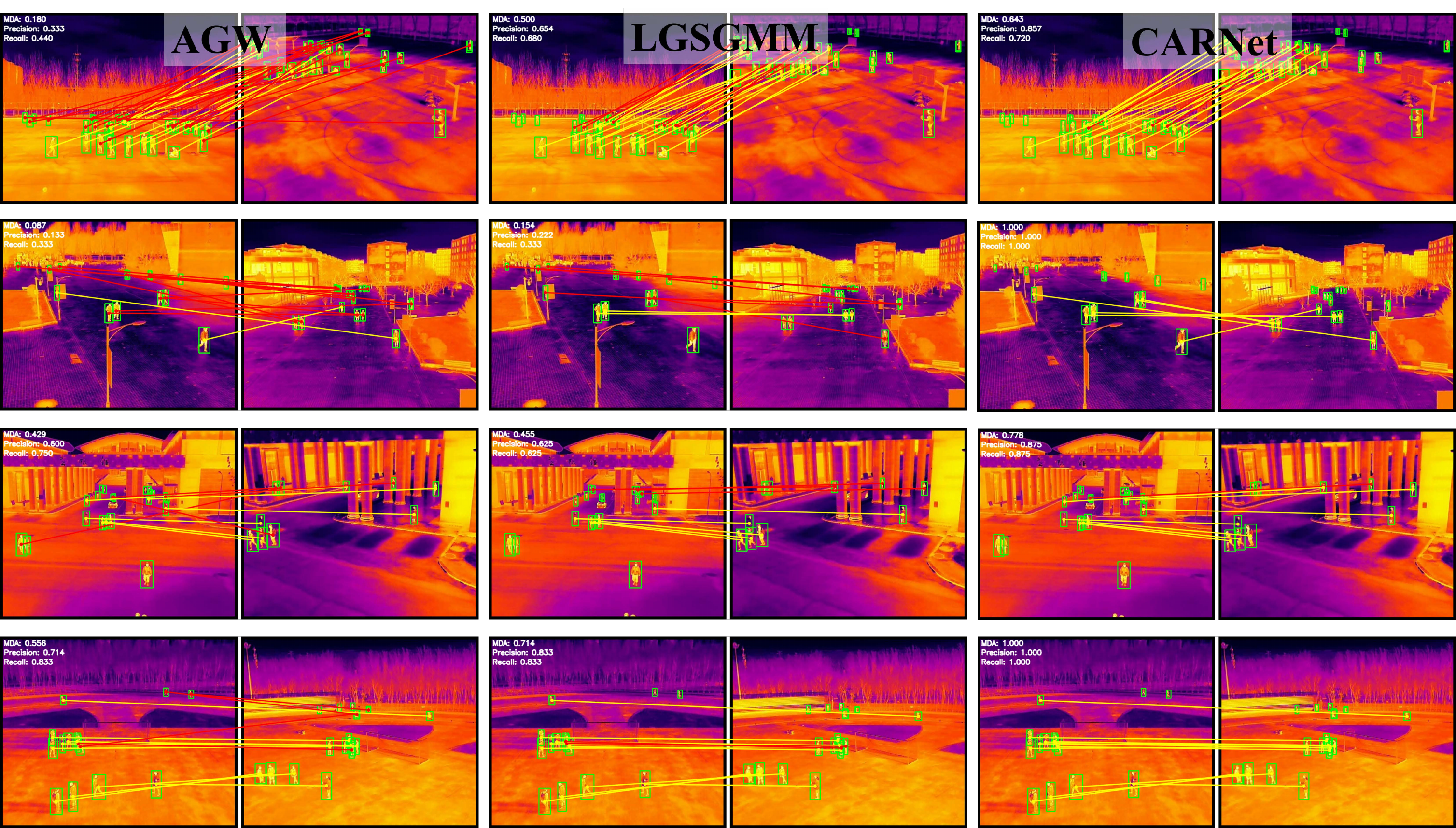}
\caption{Visualization of cross-view target association results on the infrared modality of the M3Track dataset.}
\label{m3红外}
\end{figure}

\begin{figure}[t]
	\centering
\includegraphics[width=0.9\columnwidth]{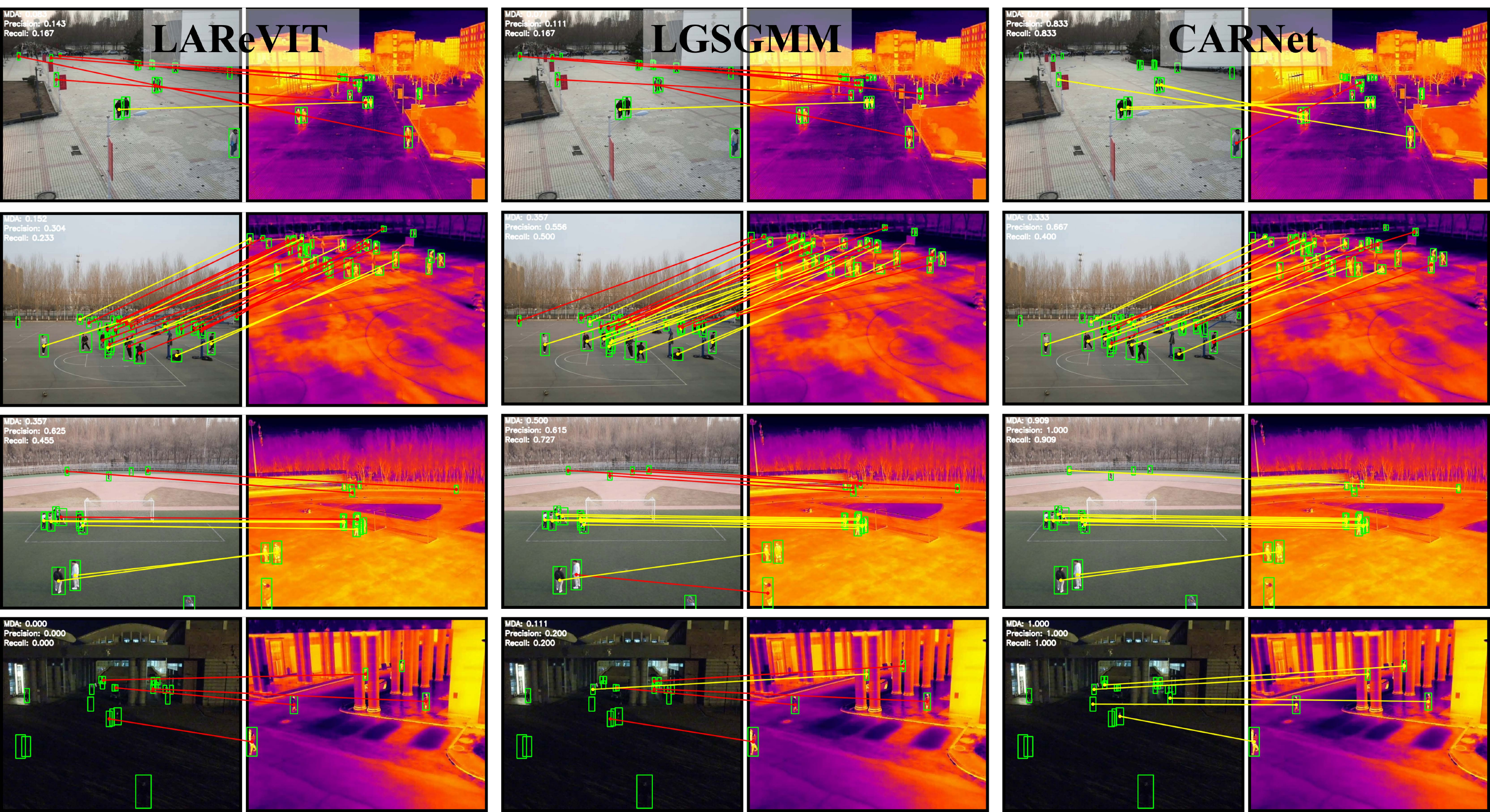}
\caption{Visualization of cross-modal target association results between visible and infrared modalities on the M3Track dataset.}
\label{m3跨膜太}
\end{figure}

\subsubsection{Comparison on M3Track}

To further evaluate the effectiveness of the proposed method in scenarios with severe non-rigid target deformation and cross-modal variations, we compare it with representative state-of-the-art cross-view object association methods on the M3Track dataset, including OSNet, CrossMOT , CT , AGW , ReIDMamba, GTA-Net, TTS, MambaGlue, and MIA-Net. 
For the cross-modal scenario, we further compare CARNet with representative cross-modal person re-identification methods, including DLM~\cite{ye2025dual}, DSAF~\cite{jiang2025dsaf}, and LAReVIT
~\cite{xu2026bit}.
The quantitative results are reported in Table~\ref{tab:m3}. As shown in the table, the proposed method achieves the best overall performance among all competing methods.

Fig.~\ref{m3可见光}, Fig.~\ref{m3红外} and Fig.~\ref{m3跨膜太}  shows qualitative association results. Appearance-based association methods, such as AGW and CT, mainly rely on target visual similarity and therefore struggle to distinguish densely distributed targets with similar appearances. LGSGMM implicitly introduces positional cues to guide attention aggregation. However, when targets are heavily occluded or observed across different modalities, inaccurate attention aggregation may introduce misleading context and reduce feature discriminability.  As shown in the figure, CARNet can still accurately associate targets under severe non-rigid deformations, dense crowds, and heavy occlusions. This robustness benefits from explicitly incorporating positional information into pairwise relation modeling, which enables more reliable association under challenging scenarios. Moreover, CARNet maintains strong performance in both same-modal and cross-modal association scenarios.


\subsection{Generalization to Cross-Camera Multi-Object Tracking}
\begin{table}[t]
	\centering
	{
		\setlength{\tabcolsep}{2pt}
		\begin{tabular}{ccccc}
			\toprule
			Dataset & Method & MOTA (\%) & IDF1 (\%) \\
			\midrule
			\multirow{8}{*}{Wildtrack}
			& KSP-DO(CVPR 2018)        & 69.6 & 73.2 \\
			& KSP-DO-track(CVPR 2018)    & 72.2 & 78.4 \\
			& GLMB-DO (TPAMI 2018)        & 70.1 & 72.5 \\
			& DMCT  (Arxiv 2020)          & 72.8 & 77.8 \\
			& DMCT Stack (Arxiv 2020)      & 74.6 & 81.9 \\
			& ReST (CVPR 2023)           & 84.9 & 86.7 \\
			& TSMMT(TCSVT 2025) & 86.4& 88.1 \\
			& \textbf{			CARNet(Ours)  }& \textbf{86.9} & \textbf{88.5} \\
			\midrule
			\multirow{8}{*}{MvMHAT}
			& Tracktor++ (ICCV 2019)     & 66.5 & 46.1 \\
			& CenterTrack (ECCV 2020)    & 63.5 & 38.1 \\
			& TraDeS (CVPR 2021)         & 69.5 & 44.9 \\
			& TrackFormer(CVPR 2022)     & 70.4 & 49.6 \\
			& DeepCC  (CVPR 2018)        & 63.9 & 44.4 \\
			& SVT    (TPAMI 2022)       & 65.4 & 47.6 \\
			& TSMMT(TCSVT 2025)  & 71.6 & 51.6 \\
			& \textbf{			CARNet(Ours) }  & \textbf{72.3} & \textbf{52.0} \\
			\bottomrule
		\end{tabular}
	}
	\caption{Comparison on two natural scenario datasets: Wildtrack and MvMHAT. The best scores are marked in bold.}
	\label{track}
\end{table}

We further evaluate CARNet on natural cross-view tracking datasets, including Wildtrack and MvMHAT. As shown in Table~\ref{track}, we compare our method with representative multi-camera tracking and association methods, including KSP-DO\cite{chavdarova2018wildtrack}, KSP-DO-track\cite{chavdarova2018wildtrack}, GLMB-DO\cite{chavdarova2018wildtrack}, DMCT\cite{you2020real}, DMCT Stack\cite{you2020real}, ReST\cite{cheng2023rest}, TSMMT\cite{wu2024temporal}, Tracktor++\cite{bergmann2019tracking}, CenterTrack\cite{zhou2020tracking}, TraDeS\cite{wu2021track}, TrackFormer\cite{meinhardt2022trackformer}, DeepCC\cite{ristani2018features}, and SVT. TSMMT uses temporal information from adjacent video frames, while CARNet only uses the current cross-view image pair without relying on preceding or subsequent frames.

Despite this more restricted input setting, CARNet achieves the best results on both datasets. On Wildtrack, CARNet obtains 86.9\% MOTA and 88.5\% IDF1. On MvMHAT, it achieves 72.3\% MOTA and 52.0\% IDF1. These results demonstrate that the proposed position-aware all-pairs relation modeling is effective for natural cross-view object association and can provide strong association performance even without temporal video cues.

\begin{table}[htbp]
	\centering
	{
		\setlength{\tabcolsep}{2pt}
		\begin{tabular}{ccccc}
			\toprule
			Method & Position & Strategy 1 & Strategy 2 & MDA (\%) \\
			\midrule
			- &  &  & $\checkmark$ & 23.19\\
			- &  & $\checkmark$ & $\checkmark$ & 31.80 \\
			- & $\checkmark$ &  &  $\checkmark$ & 33.39 \\
			CARNet & $\checkmark$ & $\checkmark$ & $\checkmark$ & 40.81 \\
			\bottomrule
		\end{tabular}
	}
	\caption{Ablation study of the proposed modules on the MDMT-CD dataset. ``Position'' denotes the Position information, ``Strategy 1'' denotes the Hard-Distinguishing Intervention, and ``Strategy 2'' denotes the Visual-Evidence Intervention.}
	\label{ablation}
\end{table}
\subsection{Ablation Study}

\begin{figure}[t]
	\centering
	\includegraphics[width=0.8\columnwidth]{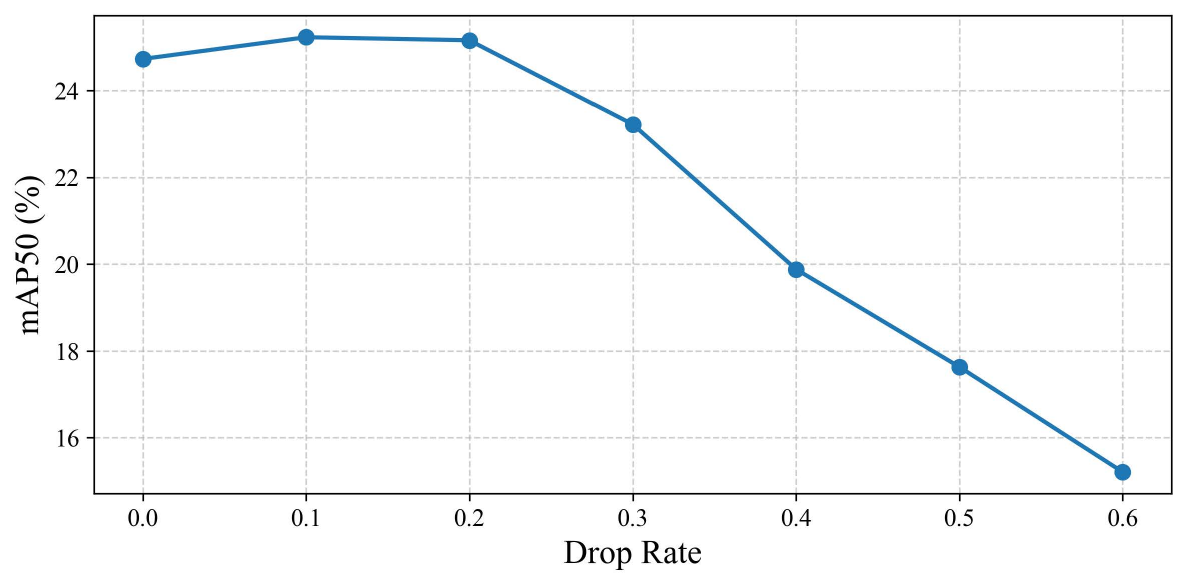}
	\caption{Effect of different Drop rates on mAP50.}
	\label{fig:drop_map50}
\end{figure}
To evaluate the effectiveness of position-aware relation modeling and the proposed intervention strategy, we conduct ablation studies on the MDMT dataset. As shown in Table~\ref{ablation}, removing either module causes a significant decrease in MDA, indicating that both components are critical for robust cross-view correspondence. Position-aware relation modeling improves the discriminability of visually similar targets by incorporating spatial cues, while the intervention strategy enhances the learning of discriminative correspondence evidence from appearance-confusing target pairs.

To investigate the effect of Visual-Evidence Perturbation Intervention, we conduct an ablation study on MDMT-CD with different dropout rates, as shown in Fig.~\ref{fig:drop_map50}. As the dropout rate increases, collaborative recognition performance first improves and then declines. This is expected: moderate perturbation prevents the model from over-relying on unstable visual dimensions and encourages stable relation representations under partial evidence loss, improving robustness to viewpoint changes, occlusion, and appearance variations. However, excessive dropout removes too much discriminative evidence, weakening reliable association and recognition.

\subsection{Paramater and Glops Analysis}

We further evaluate the scalability of CARNet by replacing the feature extraction backbone with networks of different capacities. As shown in Table~\ref{ablation33}, CARNet can be flexibly adapted to different efficiency requirements. With the same ResNet-50 backbone, CARNet uses only about two-thirds of the parameters of LGSGMM and achieves better MDA performance, demonstrating the parameter efficiency of the proposed pairwise relation framework.

When lightweight deployment is required, smaller backbones can be adopted. ResNet-18 greatly reduces the model size and computation while maintaining an acceptable MDA decrease. MobileNetV3 and ShuffleNetV2 further reduce the computational cost, but the performance drop becomes more evident. These results show that CARNet provides a flexible accuracy-efficiency trade-off for different deployment scenarios.

We further evaluate CARNet with different single-drone detectors. As shown in Table~\ref{ablation34}, CARNet consistently improves collaborative recognition across all evaluated detectors. Moreover, larger gains are achieved with weaker detectors, demonstrating the robustness and detector-agnostic nature of the proposed framework.
\begin{table}[ht]
	\centering
	{
		\setlength{\tabcolsep}{2pt}
		\begin{tabular}{ccccc}
			\toprule
			Method & Backbone & Parameters (M) & GFLOPs (G) & MDA (\%) \\
			\midrule
			\multirow{5}{*}{CARNet}
			& ResNet101 & 54.88 & 1294.33 & 41.04 \\
			& ResNet50 & 24.56 & 680.18 & 40.81 \\
			& ResNet18 & 11.83 & 304.55 & 35.93 \\
			& MobileNetV3 & 1.35 & 28.97 & 29.27 \\
			& ShuffleNetV2 & 0.86 & 18.41 & 28.22 \\
			\midrule
			LGSGMM & ResNet50 & 35.88 & 680.20 & 39.29 \\
			\bottomrule
		\end{tabular}
	}
	\caption{Ablation study of different backbones on the MDMT-CD dataset. The input image resolution is $1920\times1080$.}
	\label{ablation33}
\end{table}
\begin{table}[htbp]
	\centering
	{
		\setlength{\tabcolsep}{2pt}
		\begin{tabular}{cccc}
			\toprule
			Single Detector & mAP$_{50}$(S) (\%) & mAP$_{50}$(M) (\%) & Gain (\%) \\
			\midrule
			YOLOv3-Tiny & 18.36 & 24.15 & +5.79 \\
			YOLOv3      & 19.73 & 24.22 & +4.49 \\
			YOLOv5-s    & 22.78 & 25.03 & +2.25 \\
			YOLOv8-s   & 23.45 & 25.23 & +1.78 \\
			YOLOv12-s   & 23.64 & 25.37 & +1.73 \\
			YOLOv26-s   & 23.59 & 25.33 & +1.74 \\
			\bottomrule
		\end{tabular}
	}
	\caption{Robustness of CARNet to different single-drone detectors. }
	\label{ablation34}
\end{table}
%

\section{Conclusion}

In this paper, we proposed CARNet  and established MDMT-CD, a fine-grained benchmark for collaborative target detection. Extensive experiments demonstrate that CARNet achieves state-of-the-art performance on MDMT-CD,  M3Track, Wildtrack, and MvMHAT. These results show that the proposed method is effective for collaborative detection, cross-view target association, and cross-camera multi-object tracking, providing a strong and practical solution for multi-drone perception. However, CARNet relies on the outputs of single-drone detectors. If a target is missed by the detector, the subsequent association and fusion stages cannot recover it. Inaccurate bounding-box localization may also affect the extracted appearance features and positional representations. Future work will therefore explore the joint optimization of single-drone detection, cross-view target association, and multi-view fusion recognition.

\bibliographystyle{elsarticle-num-names}
\bibliography{ispr_references}

\end{document}